\documentclass[aps,epsf,twocolumn]{revtex4}

\usepackage{amsthm}
\usepackage{graphicx}
\usepackage{pseudocode}
\usepackage{physics}
\usepackage{bbold}

\newtheorem{theorem}{Theorem}

\begin{document}

\title{Configuration Path Control}

\author{Sergey \surname{Pankov}}
\affiliation{Harik Shazeer Labs, Palo Alto, CA 94301}

\begin{abstract}

Reinforcement learning methods often produce brittle policies -- policies that perform well during training, but generalize poorly beyond their direct training experience, thus becoming unstable under small disturbances. To address this issue, we propose a method for stabilizing a control policy in the space of configuration paths. It is applied post-training and relies purely on the data produced during training, as well as on an instantaneous control-matrix estimation. The approach is evaluated empirically on a planar bipedal walker subjected to a variety of perturbations. The control policies obtained via reinforcement learning are compared against their stabilized counterparts. Across different experiments, we find two- to four-fold increase in stability, when measured in terms of the perturbation amplitudes. We also provide a zero-dynamics interpretation of our approach. 

\end{abstract}

\maketitle

\newcommand{\T}{\mathsf T}

\section{Introduction}

The past decade has seen successful applications of deep neural networks (NN) to various machine learning tasks, such as image classification \cite{krizhevsky2012imagenet,lecun15deep}, speech recognition \cite{hinton12deep} and language translation \cite{young2018recent,singh2017machine}. In the field of reinforcement learning (RL), the employment of deep NNs as expressive function approximators has been crucial in tackling many difficult problems involving an agent interacting with its environment with the goal of maximizing its rewards. Among the notable examples of deep RL applications are: playing arcade games \cite{mnih2013playing} and challenging board games, such as Go \cite{gibney2016google}, often at super-human level; solving continuous control tasks \cite{duan2016benchmarking}, such as cart-pole and acrobot; 2D and 3D locomotion \cite{schulman2015trust,schulman2015high} and manipulation \cite{gu2017deep} tasks, including complex high-dimensional setups. In many cases, elaborate policies were learned from scratch, with little to no task-specific engineering.

However, despite numerous successes within simulated environments, there is a paucity of the real-world deployment of RL trained robots. Consider, for example, legged robots. The design of control policies of the most agile bipedal \cite{kuindersma16,gong2019feedback,shigemi2018asimo} and quadrupedal \cite{raibert2008bigdog,bledt2018cheetah,hutter2016anymal} robots appears to have little to do with RL. Among the main limiting factors, precluding the application of RL to physical robots, are high training sample complexity \cite{levine2018learning} and policy brittleness (also known as the reality gap) \cite{kober2013reinforcement,hanna2017grounded}. These deficiencies are related to the lack of generalization and overfitting. A popular way to counter them is to train on more samples and on more diverse samples. Domain randomization \cite{sadeghi2016cad2rl,tobin2017domain,james2017transferring,muratore2019assessing}, training with perturbations \cite{wang2010optimizing,venkataraman2019recovering} and adversarial training \cite{pinto2017robust,mandlekar2017adversarially} are among the methods incorporating such strategy. 

By design, the above approaches improve generalization by collecting more training samples (in a simulated environment). In this work we pursue a very different route: we investigate an approach for stabilization of RL-trained continuous control policies that does not resort to generation of additional training samples. Instead, our algorithm takes the data, produced by a policy during training with a RL algorithm, and converts it to a policy with superior stability.

The fundamental ingredient of our approach is a control policy stabilization around desired configuration paths (time-reparameterized trajectories) that can be rigorously justified in the high gain limit (HGL). We call this approach Configuration Path Control (CPC). It has some overlap with the zero dynamics (ZD) concept \cite{byrnes1991asymptotic}, which is central to the Hybrid Zero Dynamics (HZD) framework in the context of bipedal walkers \cite{westervelt2018feedback,grizzle2017virtual}. We reinterpret the derived CPC control law in the language of the HZD literature, by re-stating it in terms of the reparameterization invariant virtual constraints. 

We empirically validate the CPC method by stabilizing RL-trained policies of a bipedal walker. The RL policies are compared with their CPC-stabilized counterparts in a number of stability testing experiments, showing remarkable improvements in stability in all of the tests.

The paper is organized as follows. In Sec. \ref{cpcformulation} we derive the control law and present the CPC algorithm. Relation to virtual constraints of ZD is explained in Sec. \ref{virtconstrzerodyn}. The approach is experimentally validated in Sec. \ref{experiments}, followed by a concluding discussion in Sec. \ref{discussion}.

\section{CPC formulation}
\label{cpcformulation}

In this section we provide the underlying theory of our approach and present the CPC algorithm.

\subsection{Control law derivation}

\label{controllaw}

In this section we derive the CPC control law, by deriving and solving a trajectory reachability equation in an underactuated system.

\subsubsection{Preliminaries}
\label{prelims}

Let $q$ be the $N$-dimensional vector of configuration variables of the system. We use the dot notation for the time derivative to represent velocity $\dot q$ and acceleration $\ddot q$. Let $x \equiv [q;\dot q]$ denote the state vector. The system dynamics are governed by the equation of motion
\begin{equation}
D(q)\ddot q + H(q,\dot q) = B_{\tau}\tau,
\label{eqmot}
\end{equation}
where $D(q)$ is the inertia matrix (which is positive definite), $H(q,\dot q)$ includes the terms dependent on $q$ and $\dot q$, such as Coriolis, centrifugal and gravitational forces, $\tau$ is the $M$-dimensional vector of control variables and $B_\tau$ is the control distribution matrix. Through this section we assume the system to be fully actuated. 
A fully actuated system (full actuation requires $B_\tau$ to have rank $N$) can be forced to move with an arbitrary acceleration $u$ with the following choice of controls (this choice minimizes actuation cost $\propto \tau^\T \tau$) 
\begin{equation}
\tau_{ff}(q,\dot q, u) = B_{\tau}^{+} \left(D(q) u + H(q,\dot q) \right),
\end{equation}
which we call the feedforward control term, where $B_{\tau}^{+} = B_{\tau}^\T(B_{\tau}B_{\tau}^\T)^{-1}$ is the right pseudoinverse. The system in an initial state $x_d(0) = [q_d(0);\dot q_d(0)]$ can then in principle follow an arbitrary desired trajectory $q_d(t)$ by setting the controls to $\tau_d = \tau_{ff}(q,\dot q, u_d)$, where $u_d = \ddot q_d$. In practice, due to accumulation of errors in $q$ and $\dot q$, the system also needs a stabilizing feedback control term $\tau_{fb}$ to track $q_d(t)$.

In this paper, unless noted otherwise, we use $\Delta$ to indicate an error term, that is the deviation of a quantity from its desired value, thus $\Delta a \equiv a-a_d$ for any quantity $a$, unless specified otherwise. We define the control matrix $B(q) \equiv D^{-1}(q)B_\tau$, so 
\begin{equation}
  \Delta \ddot q = B(q)\Delta\tau,
\label{delubdeltau}
\end{equation}
where $\Delta \ddot q = \ddot q - \ddot q_d$ and $ \Delta\tau = \tau - \tau_{ff}(q,\dot q, \ddot q_d)$. Let us define the problem of tracking $q_d(t)$, within the optimal control framework \cite{anderson2007optimal}, by penalizing $\Delta q$ and $\Delta u$ with quadratic cost terms $c_q{\Delta q}^\T \Delta q$ and $c_u{\Delta u}^\T \Delta u$. The optimal control solution for the feedback term is a linear controller
\begin{equation}
\tau_{fb} = - B^{-1}(q) K \Delta x, \quad K = \left[\frac{k_p}{\varepsilon^2}, \frac{k_d}{\varepsilon}\right] \otimes I_N,
\label{taufbfullact}
\end{equation}
where $\Delta x = [\Delta q, \Delta \dot{q}]$, $\varepsilon = (c_u/c_q)^{\frac{1}{4}}$, $k_p$ and $k_d$ are $\mathcal{O}(1)$ and $I_N$ denotes $N \cross N$ identity matrix. We see (from Eqs.(\ref{delubdeltau},\ref{taufbfullact})) that with the control law $\tau = \tau_{ff} + \tau_{fb}$, $\Delta q$ follows the dynamics of a near-critically damped oscillator:
\begin{equation}
\Delta \ddot q + \frac{k_d}{\varepsilon} \Delta \dot q + \frac{k_p}{\varepsilon^2} \Delta q = 0
\label{errordyn}
\end{equation}
It implies that the error terms vanish exponentially fast on the time scale $\mathcal{O}(\varepsilon)$, while $\Delta \ddot q = \mathcal{O}(\Delta \tau)$, $\Delta \dot q = \mathcal{O}(\Delta \tau \varepsilon)$ and $\Delta q = \mathcal{O}(\Delta \tau \varepsilon^2)$. Note, we are using $\mathcal{O}(a)$ notation as synonymous to ``of the order of $c a$'', where $c$ is a (dimensionful, in general) constant of order one. We see that in the HGL $\varepsilon \to 0$, the terms $\Delta q$ and $\Delta \dot q$ can be ignored relative to $\Delta \ddot q$ and $\Delta\tau$ to leading order in $\varepsilon$.

We define the configuration path of a trajectory $q(t)$ as the set of all configurations visited by $q(t)$. Thus, two different trajectories $q(t)$ and $q'(t)$, related by a time reparameterization $q(t'(t))=q'(t)$, map to the same configuration path. In that case we may say that $q'(t)$ is associated with the path of $q(t)$, and vice versa. In this paper, we use interchangeably the terms: ``configuration path'' and ``path'', as well as, ``desired'' and ``target'' trajectory.

\subsubsection{Motivations of CPC formulation}
\label{motivations}

Unlike the case of full actuation in the previous section, many important control problems, such as walking on point feet \cite{westervelt2018feedback}, are essentially underactuated, which makes them much more challenging. Motivated by the desire to make the underactuated control problem more tractable, while preserving its safety critical aspects, we will resort to stabilization around a time-independent configuration path, instead of stabilization around a time-dependent trajectory. Giving up the time dependence makes the control problem easier, without affecting the robot's safety metrics, provided they are defined in terms of the configuration variables. Indeed, a collision indicator -- be it collisions with obstacles, the ground or between the robot parts -- is typically defined in terms of configuration variables alone.

Another crucial motivation stems from our aim for stabilization of model-free RL policies, in which we at most can afford a quasi-instantaneous estimation of the local dynamics based on the $\mathcal{O}(N)$ most recent points along the current trajectory. Without possibility of longer time planning, we will be assuming short stabilization time scale $\varepsilon$. In other words, we derive the CPC control law under the assumption of HGL, in which case we only need to retain the leading terms in $\varepsilon$. When we omit sub-leading terms in an equation, we will either explicitly indicate its precision by adding a precision term to the right hand side of the equation or use the approximately equal sign.
See Sec. \ref{virtconstrzerodyn} for a discussion on the satisfiability of HGL in specific robotic systems.

\subsubsection{Reachability equation}
\label{reachability}

We define trajectory reachability in the HGL, specified by the scale $\varepsilon$, as follows: the system can reach a trajectory $x_d(t)$ from a given state if it can be driven to $x_d(t)$ in time $t = \mathcal{O}(\varepsilon)$. A fully actuated system can move with an arbitrary acceleration, and therefore, can reach any trajectory. An underactuated system, on the contrary, can only reach a subset of trajectories. In this section we derive the trajectory reachability equation and use it to analyze the configuration path reachability in the underactuated case. 

Let $q_d(t)$ be the system's trajectory under some choice of controls $\tau_d(t)$ starting at $x_d(0)$. Similarly for $q(t)$, $\tau(t)$ and $x(0)$. We subtract from Eq.(\ref{eqmot}) for $q$ the same equation for $q_d$ to find 
\begin{equation}
  \Delta \ddot q = B(q)\Delta\tau 
  +\mathcal{O}(\Delta\tau\varepsilon),
\label{delubdeltauhighgainlim}
\end{equation}
(which is only superficially similar to Eq.(\ref{delubdeltau}), since $\Delta\tau$ has now a different meaning). Without loss of generality we assume the independence of the first $M$ rows of the control matrix (i.e. $B(q)$ is full rank), while underactuation implies $N > M$. Therefore, we can write $B(q) = [B_\chi(q);B_\psi(q)]$, where $B_\chi(q)$ is invertible. Accordingly, we write $q = [\chi;\psi]$, where $\dim(\chi) = M$. We call $\chi$ and $\psi$ the controlled and free coordinates respectively. Writing Eq.(\ref{delubdeltauhighgainlim}) as a system of two equations, $\Delta{\ddot \chi} \approx B_{\chi}(q) \Delta\tau$ and $\Delta{\ddot \psi} \approx B_{\psi}(q) \Delta\tau$, then using the invertibility of $B_{\chi}(q)$ in the first equation to eliminate $\Delta\tau$ in the second equation, we obtain 
\begin{equation}
  b^\T(q) \Delta{\ddot q} = 0 
  +\mathcal{O}(\Delta\tau\varepsilon),
  \label{bddq}
\end{equation}
where
\begin{equation}
  b(q) = [B_{\psi}(q)B_{\chi}^{-1}(q),-I_{N-M}]^\T,
\end{equation}
so the set of possible accelerations $\Delta\ddot q$ is the null-space of $b^\T(q)$.

Let us define $b\equiv b(q(0))$, and similarly for $B$, $B_\chi$ and $B_\psi$, so $b = [B_{\psi}B_{\chi}^{-1},-I_{N-M}]^\T$. Note that $b = b(q(\varepsilon)) + \mathcal{O}(\varepsilon)$. If $q_d$ is reachable, the system can be driven to $q_d$, so $\Delta q(t) = 0$ for $t > t_c = \mathcal{O}(\varepsilon)$. In that case, integrating Eq.(\ref{bddq}) twice from $0$ to $t > t_c$ we find $b^\T \left( \Delta q(0) + t \Delta\dot q(0) \right) \approx 0$. The equality must hold for any $t > t_c$, leading to the trajectory reachability equation 
\begin{equation}
  ({\rm diag}{(1,\varepsilon)} \otimes b^\T) \Delta x(0) = 0 
  +\mathcal{O}(\Delta\tau\varepsilon^3).
  \label{reacheq}
\end{equation}
Note, the derivation of the reachability equation implies that a target trajectory $q_d(t)$ in Eq.(\ref{reacheq}) must be realizable under the system dynamics by some choice of controls $\tau_d(t)$. The trajectory reachability equation can also be used to determine configuration path reachability. We say the configuration path of $q_d(t)$ is reachable if there exists a reachable trajectory $q'_d(t)$, related to $q_d(t)$ by time reparameterization. Below we will solve the path reachability to linear order in time reparameterization.

Before we proceed, let us introduce some shorthand notations. We will optionally use a superscript in the error term, to differentiate between different target trajectories, so $\Delta^c a \equiv a-a_d^c$. Unless specified otherwise, $\Delta' q \equiv q-q'_d$, where $q'_d$ is defined above. We reserve $0$ superscript to indicate a value at $t=0$, so $a^0 \equiv a(t=0)$. We also define $\bar a \equiv b^\T a$. We may use $\mathcal{O}(a_1, a_2, ... , a_n)$ notation to represent $\sum_{i=1}^{i=n} \mathcal{O}(a_i)$. 

Let $q'_d(t) = q_d(t'(t))$ and $t_0$ be the root of $t'(t) = 0$. The linear order expansion of $t'(t)$ about $t'=0$ is $t'(t) \approx (t-t_0)/s$, where $s = (\partial t'(t = t_0)/\partial t)^{-1}$. Therefore, $q'_d(t) = q_d^0 + \dot q_d^0 (t-t_0)/s + \mathcal{O}(t^2+t_0^2)$. Substituting $q'_d$ in the reachability equation Eq.(\ref{reacheq}) we obtain a linear system of $2(N-M)$ equations on $t_0$ and $s$. In the case of one degree of underactuation, $N-M = 1$, it is satisfied exactly by
\begin{equation}
  t_0 = \frac{{\bar q}_d^0-{\bar q}^0}{\dot{\bar q}^0}, \quad
  s = \frac{\dot{\bar q}_d^0}{\dot{\bar q}^0}.
\label{t0s}
\end{equation}
The considered linear approximation is justified provided $s = \mathcal{O}(1)$ and $|t_0| \ll \mathcal{O}(\dot{\bar q}^0)$. In practice, (in the discrete time domain and with an abundance of target points), provided a system stays close to its target trajectory, these conditions are either always satisfied or only briefly violated (when the sign of $\dot{\bar q}^0$ flips). Therefore, we limit our CPC algorithm implementation, as well as the theoretical investigation in the rest of the paper, to the above linear analysis, which assumes $\dot{\bar q}^0 \ne 0$ and $\dot{\bar q}_d^0 \ne 0$.

Let us define the renormalized target trajectory $q_d^r$ as the linear part of a reachable trajectory $q'_d$, corresponding to the path of $q_d$:
\begin{equation}
  q_d^r(t) = q_d^0 + \frac{\dot q_d^0}{s} (t-t_0),
  \label{qrd}
\end{equation}
where $t_0$ and $s$ are given by Eq.(\ref{t0s}). Our analysis implies that for one degree of underactuation, to linear order in $t$ and $t_0$, the renormalized target trajectory $q_d^r$ depends on $q$ and the configuration path of $q_d$, but not on the time parameterization of $q_d$. Indeed, from Eqs.(\ref{t0s},\ref{qrd}), one finds to linear order:
\begin{equation}
  {\bar q}_d^r(t) = \bar q(t).
  \label{bqdrbq}
\end{equation}
Our analysis also implies that both $q_d^r$ and the path of $q_d$ are reachable with precision $\mathcal{O}(\Delta\tau\varepsilon^3, \varepsilon^2 + t_0^2)$, for $N-M=1$. It is therefore desirable to select target points $x_d$ with $t_0 \approx 0$. To that end, we will be evaluating the quality of target points using a certain score function measuring how close $t_0$ is to $0$, among other criteria. Note, for $N-M>1$, the reachability equation can only be solved approximately for a general $q_d$. In that case we can include a loss term in the score function, quantifying the error of Eq.(\ref{reacheq}). The reparameterization parameters $t_0$ and $s$ can then be determined by minimizing the score function. Assuming the loss term in the form $f_0(\lVert \Delta' \bar q(0)\rVert) + f_1(\lVert \Delta' \dot{\bar q}(0)\rVert)$, with the only requirement for $f_i(y)$ to have a global minimum at $y=0$, we find a generalization of Eq.(\ref{t0s}) to $N-M > 1$:
\begin{equation}
  t_0 = \frac{{\dot{\bar q}_d^0}^\T \! \left({\bar q}_d^0-{\bar q}^0\right)}{{\dot{\bar q}_d^0}^\T\dot{\bar q}^0}, \quad
  s = \frac{{\dot{\bar q}_d^0}^\T\dot{\bar q}_d^0}{{\dot{\bar q}_d^0}^\T\dot{\bar q}^0}.
  \label{t0shigher}
\end{equation}
A more detailed consideration of the higher degree of underactuation case $N-M>1$ is beyond the scope of this paper.

\subsubsection{CPC control law}

Let $x_\chi \equiv [\chi;\dot\chi]$ and $\chi_d^r(t) = \chi_d^0 + \dot \chi_d^0 (t-t_0)/s$, where $t_0$ and $s$ are given by Eq.(\ref{t0shigher}), (or by Eq.(\ref{t0s}) for the special case of $N-M=1$), for the current state $x = [q^0,\dot q^0]$ and a target state $x_d = [q_d^0,\dot q_d^0]$. We define the CPC control law as (cf. Eq.(\ref{taufbfullact})):
\begin{equation}
  \tau_{CPC}(t) = \tau_d - B_{\chi}^{-1} K \Delta^r x_\chi(t), \quad K = \left[\frac{k_p}{\varepsilon^2}, \frac{k_d}{\varepsilon}\right] \otimes I_M.
  \label{taucpc}
\end{equation}
If $\Delta^r x$ satisfies the reachability equation, the CPC control law will drive the system to the configuration path of $q_d$ despite underactuation. Moreover, the control law is invariant with respect to the choice of the controlled coordinates. These two statements are formalized and proven (see App. \ref{proofs} for proofs) in the following theorems.

\begin{theorem}[ (path convergence theorem)]
\label{convergence}
Assume $\dot{\bar q}^0 \ne 0$ and $\dot{\bar q}_d^0 \ne 0$. Let $\tau_{CPC}(t)$ be defined by Eq.(\ref{taucpc}). If $\Delta^r x(0)$ satisfies the reachability equation Eq.(\ref{reacheq}), then under $\tau_{CPC}(t)$, within the HGL $\varepsilon \to 0$, the error term $\Delta' q(t)$ reduces to $\mathcal{O}(\varepsilon \Delta' q(0), \varepsilon^2+t_0^2)$ on the time scale $\mathcal{O}(\varepsilon)$.
\end{theorem}

\begin{theorem}[ (reparameterization invariance theorem)]
\label{reparameterization}
Let all the conditions of Theorem \ref{convergence} hold. Let $\tilde q = [\tilde\chi,\tilde\psi] = C q$, $\tilde B = [\tilde B_\chi,\tilde B_\psi] = C B$, where $C$ is a linear invertible transformation and $\tilde B_\chi$ is invertible. Other quantities defined through $q$ and $B$ change correspondingly, e.g. $x \to \tilde x = [\tilde q; \dot{\tilde q}]$ and $b \to \tilde b = [\tilde B_{\psi} \tilde B_{\chi}^{-1},-I_{N-M}]^\T$. Let $\tilde\tau_{CPC}(t)$ be defined by Eq.(\ref{t0shigher}) and Eq.(\ref{taucpc}), with $q$, $x$, $B$, $b$ replaced by the corresponding transformed quantities $\tilde q$, $\tilde x$, $\tilde B$, $\tilde b$. Then, within the HGL, $\tilde\tau_{CPC}(t) = \tau_{CPC}(t)$.
\end{theorem}

The reparameterization invariance theorem and its proof may appear obvious. However, this invariance is specific to our formulation and does not generally hold for the ZD control law designs proposed in the HZD literature, which we denote $\tau_{ZD}$. Rather, $\tau_{ZD}$ is formulated in terms of virtual constraints \cite{westervelt2018feedback,grizzle2017virtual}, requiring manual selection of a gait phasing parameter and ensuring that it is monotonically increasing with time, for the type of motion under consideration. Generally, $\tau_{ZD}$ is sensitive to the choice of free coordinates. In our approach, on the contrary, the selection of free coordinates has no bearing on $\tau_{CPC}$, as long as $B_\chi$ is full rank. Due to this invariance, the definition of $\tau_{CPC}$ does not requires a foresight into the type of motion considered, (unlike $\tau_{ZD}$), and is therefore suitable for stabilization of a generic RL-learned policy. A rigorous connection between our approach and HZD is provided in Sec. \ref{virtconstrzerodyn}, where we express the CPC control law in the language of virtual constraints.

\subsection{Selection of target points}
\label{tpselection}

In the previous section we have introduced the CPC control law (see Eq.(\ref{taucpc})), that specifies controls $\tau_{CPC}$ for a given target point $x_d$, that drives the system to a configuration path containing $q_d^0$. While $\tau_{CPC}$ is formally defined for any $x_d$, the best possible target points should be selected for computing $\tau_{CPC}$, because the quality of the adopted approximations strongly depends on the choice of $x_d$ as we explain below.

Firstly, the error term in the statement of Theorem \ref{convergence} reduces to $\mathcal{O}(t_0^2)$ in the limit $\varepsilon/t_0 \to 0$. Therefore, one should select those target points that minimize $t_0^2$. Secondly, the feedforward term $\tau_d^{ZD}$ depends on $\dot q$ (see Eq.(\ref{taudzd})), while in our application of CPC the feedforward term $\tau_d$ is taken from available RL samples. Therefore, one should generally expect $\tau_d\approx\tau_d^{ZD}$ only for $s\approx 1$. Thus, one should typically select target points with $s\approx 1$.

The path tracking precision is not the only criterion one may need to consider in the context of RL policy stabilization. In the tasks with stochastic environments or variable goals one also faces the problem of selecting the best path out of multiple candidates of generally unequal value, which is encoded in the RL value function. Therefore, the value function must be incorporated in the process of the target point selection. The selection procedure that addresses both issues is presented in the next two subsections.

\subsubsection{Selection by time reparameterization parameters with efficient ball-tree search}
\label{balltreesearch}

We assume that the training data of an RL algorithm is stored as a set $\mathbf X_d$ of data points $(x_t,\tau_t,G_t)$, representing an instance of the system's state $x$, controls $\tau$ and return $G$ (total future discounted reward, see Sec. \ref{expectedreturn}) recorded at time $t$. For brevity, we will often use the state notation $x$ alone, implying an access to the corresponding $\tau$ and $G$.

Generally, a suitable target point $x_d\in\mathbf X_d$ is selected based on the proximity of $(t_0,s)$ to $(0,s_g)$ (where usually, but not always, $s_g = 1$) and on the maximization of the expected return. We found it convenient to separate the optimization over $(t_0,s)$ from the return optimization. The main reason is that the loss function for $t_0$ and $s$ can be very efficiently optimized by storing $\mathbf X_d$ in a ball-tree search data structure, as we explain below.

Let us introduce a proximity loss 
\begin{equation}
L^{prox}(t_0, s) = (\omega t_0)^2 + (s-s_g)^2 ,
\label{proxloss}
\end{equation}
where $\omega$ balances the relative importance of $|t_0|$ versus $|s-s_g|$ smallness. The set $\mathbf X_d$, stored on a ball-tree, can be searched for best performing points, that minimize the proximity loss, by a standard branch and bound method. The central element of such a method is the bounding of the optimization function on a given tree node. The maximum efficiency is reached when the bounds are tight. 
The tight bounds can indeed be computed in closed form, as shown in App. \ref{bounds}.

\subsubsection{Selection by expected return}
\label{expectedreturn}

Depending on the problem setup, different target points may vary greatly in terms of how valuable it is for the system to visit them. The cost of getting to those points from the current state may vary greatly as well. We will quantify the utility of choosing $x_d$ as a target point in the current state $x_0$ by estimating the corresponding expected return. 

In a discrete time formulation, we define return $G_t$ at time $t$ as
\begin{equation}
G_t = \sum_{i=0}^{\infty}\gamma^i r(x_{t+i},\tau_{t+i}),
\label{return}
\end{equation}
where $r(x,\tau)$ is the reward function and $\gamma$ is the discount factor. 
In RL the goal is to maximize the expected return. The value function $V(x)$ is defined as the expected return of the system in state $x$:
\begin{equation}
V(x) = \mathbb{E}_{|x_0=x}[G_0],
\label{valfunc}
\end{equation}
The expectation is taken over trajectories $x_0\tau_0 x_1\tau_1 ...$, sampled under a given policy and the problem dynamics, starting from $x_0=x$. For simplicity, we will treat both the policy and environment as deterministic.

Let a discrete time step $\Delta t$ correspond to incrementing $i$ by $1$ in Eq.\ref{return}. Consider $\gamma$ and $r(x,\tau)$ in the form:
\begin{equation}
\gamma = 1 - \frac{\Delta t}{T_{\gamma}},
\label{gamma}
\end{equation}
\begin{equation}
r(x,\tau) = \frac{\Delta t}{T_{\gamma}} \left(\tau^\T C_{\tau}\tau +r(x)\right),
\quad C_\tau=C_\tau^\T\le 0.
\label{rxtau}
\end{equation}
where the parameter $T_{\gamma}$ sets the time scale of the reward discounting. This choice ensures that $V(x)$ is well behaved in the continuous time limit $\Delta t \to 0$, in which
\begin{equation}
V(x)=T_{\gamma}^{-1}\int_0^{+\infty}\left(\tau(t)^\T C_{\tau}\tau(t) +r(x(t))\right)e^{-\frac{t}{T_{\gamma}}}dt .
\label{value}
\end{equation}
We use this equation to estimate (in the HGL, and assuming $(t_0,s) \approx (0,1)$) the value function $V_{x_d}(x_0)$ of the system in state $x(0) = x_0$ that reaches the target trajectory $q_d$ at $x_d^r(0)$ (we call it stage I) and then follows it under the CPC control law (we call it stage II). The value function has two corresponding contributions:
\begin{equation}
  V_{x_d}(x_0) = V_I + V_{II},
  \label{vxd}
\end{equation}
where $V_I$ is dominated by the work penalty during the transition from $x(0)$ to $x_d^r(0)$:
\begin{equation}
  V_I \approx T_{\gamma}^{-1} \left(2\tau_d^\T C_{\tau}\int_0^{+\infty}\Delta\tau(t)dt 
  + \int_0^{+\infty}\Delta\tau(t)^\T C_{\tau}\Delta\tau(t)dt \right) ,
  \label{vi}
\end{equation}
and
\begin{equation}
  V_{II} \approx V(x_d^r(0)) \approx V(x_d)+\frac{t_0}{T_{\gamma}}\left(\tau_d^\T C_{\tau}\tau_d +r(x_d)-V(x_d)\right) .
  \label{vii}
\end{equation}
In this section we omit CPC subscript for brevity, so above $\Delta\tau \equiv \Delta\tau_{CPC}$. In the HGL that we consider, the integral in Eq.(\ref{vi}) can be evaluated exactly, as $\Delta\tau(t) = - B_{\chi}^{-1} K \Delta^r x_\chi(t)$, where
\begin{equation}
  \Delta^r x_\chi(t) = e^{Ft} \Delta^r x_\chi(0), \qquad F = 
  \begin{bmatrix}
    0 & 1 \\
    -\frac{k_p}{\varepsilon^2} & -\frac{k_d}{\varepsilon}
  \end{bmatrix} \otimes I_M .
\end{equation}
Below, we present the result for a critically damped controller case, $k_d = 2\sqrt{k_p}$, which we adopt through the rest of the paper. Denoting $\kappa = \sqrt{k_p}/\varepsilon$, we find 
\begin{multline}
  V_I = -\frac{2}{T_{\gamma}}\tau_d^\T C_{\tau} Z_1^\T \Delta^r x_\chi^0 \\
  +\frac{\kappa}{4T_{\gamma}} {\Delta^r x_\chi^0}^\T  
  \left( Z_1 C_{\tau} Z_1^\T + Z_2 C_{\tau} Z_2^\T \right) \Delta^r x_\chi^0,
  \label{vi1}
\end{multline}
where
\begin{equation}
  Z_i = \left(z_i \otimes I_M \right) {B_{\chi}^{-1}}^\T ,
  \quad z_1 = 
  \begin{bmatrix}
    0 \\ 
    1
  \end{bmatrix} ,
  \quad z_2 = 
  \begin{bmatrix}
    \kappa \\ 
    2
  \end{bmatrix} .
  \label{zi}
\end{equation}
Eqs.(\ref{vxd},\ref{vii},\ref{vi1},\ref{zi}) provide a closed-form expression for estimating $V_{x_d}(x_0)$. Note Eq.(\ref{vii}) contains $V(x_d)$, which in this work is simply approximated by a recorded return $V(x_d) \approx G_d$ from the corresponding triplet $(x_d, \tau_d, G_d)$.  

\subsection{Algorithm}

\vspace{1em}
\noindent\fbox{
\begin{minipage}{\dimexpr\linewidth-3\fboxsep-3\fboxrule\relax}
\begin{pseudocode}[plain]{cpcloop}{x_0,B,\mathbf X_d,\omega,s_g,n_d,k_0,\tau_c,k_c}
  \mathbf Y \GETS \CALL{candidates}{x_0,B,\mathbf X_d,\omega,s_g,n_d} \\
  k \GETS k_0 \\
  \REPEAT
  y^* \GETS \operatorname*{arg\,min}_{y \in \mathbf Y} \CALL{cost}{y,k,x_0,B} \\
  \tau \GETS \CALL{controls}{y^*,k,x_0,B} \\
  k \GETS k/2
  \UNTIL \lVert\tau\rVert < \tau_c \OR k<k_c \\
  \RETURN \tau
  \label{cpcloop}
\end{pseudocode}
\vspace{-1em}
\end{minipage}
}
\vspace{1em}

We are now in a position to present the algorithm that combines all the necessary ingredients derived in Secs. \ref{controllaw} and \ref{tpselection} . The function $\CALL{cpcloop}{}$, shown in  Alg. \ref{cpcloop}, is called at every cycle of the robot controller's loop. It is supplied with the current state $x_0$ and the control matrix estimate $B$, (in addition to a number of static parameters). It then selects the best possible target point and returns corresponding CPC control values $\tau_{CPC}$.

The selection of the best target point $x_d\in\mathbf X_d$ is done in two stages. In the first stage, the proximity loss $L^{prox}$ (see Eq.(\ref{proxloss})) is optimized to select $n_d$ best performing target point candidates. This is done by the function $\CALL{candidates}{}$ that returns a set of candidates $\mathbf Y$ containing tuples $y=(x_d^0,t_0,s)$, with $t_0$ and $s$ computed for corresponding $x_d^0 \in \mathbf X_d$ and the current state $x_0$, (see Eq.(\ref{t0s})). The function can be implemented efficiently  as described in Sec. \ref{balltreesearch}. In the second stage, out of $n_d$ preselected candidates, the best $x_d^0$ is selected based on $V_{x_d}$. To that end, we use the function $\CALL{cost}{}$ which returns $-V_{x_d}(x_0)$, (see Eq.(\ref{vxd}) and Eqs.(\ref{vii},\ref{vi1},\ref{zi})). Note that in Alg. \ref{cpcloop}, $k = \kappa^2$.

The controls $\tau$ are computed by the function $\CALL{controls}{}$ that returns $\tau_{CPC}(0)$, (see Eq.(\ref{taucpc})). In the second stage we also verify that the computed controls stay within certain bounds, so not to exceed reasonable limits of actuation. Starting from $k=k_0$, if $\lVert\tau\rVert$ exceeds $\tau_c$ we keep reducing $k$ (and recomputing $\tau$) until either $\lVert\tau\rVert$ falls within the bounds or the gain becomes unreasonably small $k \le k_c$.

All functions in Alg. \ref{cpcloop} rely on the control matrix $B$, which in this work is estimated on-line. While a higher accuracy of $B$ estimation may be beneficial for the algorithm's performance, it was not critical in our experiments. 
The control matrix was crudely estimated from Eq.(\ref{delubdeltauhighgainlim}) using a standard linear regression technique. Namely, $B$ is the least-square error solution over the last $n$ data points $(\tau^i,u^i)$, $i=1..n$, along the current trajectory, with $\tau_d$ and $u_d$ set to $0$. We choose $n$ to be slightly larger than $N$.

\section{Relation to ZD virtual constraints}
\label{virtconstrzerodyn}

We briefly overview the ZD approach \cite{westervelt2018feedback,grizzle2017virtual}. The ZD are a restriction of the full dynamics Eq.(\ref{eqmot}) to a submanifold of the configuration space manifold, defined by means of a virtual constraint
\begin{equation}
  y = h(q) \equiv \chi -h_d(\theta(q)),
  \label{virtconstr}
\end{equation}
where $y$ is the system's output that needs to be driven to zero by a ZD controller, and $h_d$ is the desired value of the controlled variable $\chi$ at the gait phasing parameter $\theta$. In this section we assume $N-M = 1$ and a scalar $\theta(q)$. For the constraint to represent the configuration path of $q_d$, one needs $h_d(\theta(q_d)) = \chi_d$ and $\theta$ to be monotonic along the trajectory $q_d$. Differentiating $h(q)$ twice with respect to time and using Eq.(\ref{eqmot}) to eliminate $\ddot q$ in the result, one finds for $\ddot y$:
\begin{equation}
  \ddot y = A(q) \left(\tau - \tau_d^{ZD}(q,\dot q)\right),
\end{equation}
where
\begin{equation}
  A(q) = \frac{\partial h(q)}{\partial q} B(q)
\end{equation}
and
\begin{multline}
  \tau_d^{ZD}(q,\dot q) = A^{-1}(q) \\
  \times \left(
  \frac{\partial h(q)}{\partial q} D^{-1}(q) H(q,\dot q)
  - \frac{\partial}{\partial q}\left( \frac{\partial h(q)}{\partial q} \dot q\right)\dot q
  \right).
  \label{taudzd}
\end{multline}
We define the ZD controller
\begin{equation}
  \tau_{ZD} = \tau_d^{ZD}(q,\dot q) - A^{-1}(q) \left(\frac{k_p}{\varepsilon^2} y + \frac{k_d}{\varepsilon} \dot y\right),
  \label{tauzd}
\end{equation}
which drives the system to $y=0$ submanifold, as
\begin{equation}
  \ddot y + \frac{k_d}{\varepsilon} \dot y + \frac{k_p}{\varepsilon^2} y = 0 .
\end{equation}

Let $h$ constrain the system to the path of $q_d$. Let a trajectory $q_d^{ZD}$ be associated with the path of $q_d$ and satisfy $\theta(q_d^{ZD}) = \theta(q)$. Due to the monotonicity of the gait phasing parameter, $q_d^{ZD}$ is unique for a given $q$. From the definition of $h$ and $q_d^{ZD}$ it follows $h = \chi - \chi_d^{ZD} = \Delta^{ZD} \chi$. As is common practice \cite{westervelt2018feedback,grizzle2017virtual}, we assume $\theta$ in the form $\theta(q) = c^\T q$, where $c$ is constant. 
We can now formulate a correspondence between the CPC and ZD control laws, (proven in App. \ref{correspondenceproof}):
\begin{theorem}[ (CPC-ZD correspondence theorem)]
\label{correspondence}
Let $\Delta\tau_{CPC} \equiv \tau_{CPC}-\tau_d$ and $\Delta\tau_{ZD} \equiv \tau_{ZD}-\tau_d^{ZD}$ be the CPC and ZD feedback control terms respectively. The feedback terms coincide in the HGL, $\Delta\tau_{CPC} \to \Delta\tau_{ZD}$ as $\varepsilon \to 0$, provided $c \propto b$.
\end{theorem}

The virtual constraints, considered thus far, are holonomic (HVC): they depend purely on the configuration variables and not the velocities. This  can be regarded as a drawback, for example, for a bipedal walker controller. Indeed, a HVC-based controller cannot adjust the walker's step size in response to a change in it's velocity, which should be helpful during push recovery. Recent development of HZD based on non-holonomic virtual constraints (NHVC) \cite{griffin2017nonholonomic,horn2018hybrid} addresses this issue, greatly improving stability under disturbances. Specifically, the velocities enter the constraints via the momentum conjugate of the unactuated degrees of freedom, rendering them relative degree two NHVCs. The invariance of ZD under impacts can be achieved by a dynamic variable reset \cite{griffin2017nonholonomic} or by a proper parameterization of NHVCs \cite{horn2018hybrid}. The approach has been validated both in simulation \cite{griffin2017nonholonomic,horn2018hybrid} and on a physical robot \cite{griffin2017nonholonomic}.

CPC also exhibits step-size adaptation under changing velocity, as can be clearly observed (see a demo video, Sec. \ref{experiments}) under strong disturbances, but it is achieved by different means. While for a single target trajectory CPC is similar to ZD under HVC (as follows from Theorem \ref{reparameterization}), the effective velocity adaptation arises thanks to the target point selection procedure, that favors trajectories with $s\approx 1$ that are picked (using efficient search) from a multitude of possibilities. 

We view HGL primarily as a methodological approach that allows us to make a number of rigorous statements, proven in the theorems. In practical terms, the proofs also provide precision terms, quantifying the degree of approximation. The limit $\varepsilon\to 0$ is an idealization, that is neither necessary nor possible to attain in a real system. The parameter $\epsilon$ regulates the stiffness of the feedback controller. In practice, it is limited by various factors, such as the controller bandwidth, torque limits, ground friction and so on. In our numerical experiments in Sec. \ref{experiments}, the initial stiffness is well below the bandwidth limitation (determined by the time discretization), and is adaptively attenuated to stay within reasonable torque limits, see Alg.\ref{cpcloop}.

\section{Experiments}
\label{experiments}

While our approach is very general, in this work we test it on two distinct stability-critical tasks: 1) bipedal walking under strong disturbances, 2) acrobot balancing derived purely from failure examples. A demo video \footnote{https://youtu.be/6OG8x-iSWAE} of the walker stability experiments is available online.

\subsection{Bipedal walker}

\subsubsection{Stability testing procedure}

Our goal in this work is to evaluate policy stability in a way that is reflective of the policy transferability to a real world setting. The mean return, which is often regarded as measuring the quality of a RL-trained policy, including the bipedal walking policies \cite{schulman2015trust,schulman2015high,duan2016benchmarking}, is a poor indicator of such stability. Indeed, even though a RL reward usually penalizes policy failures (e.g. falls), the return contains no information about the policy's performance outside of the training distribution, which can be quite narrow. Similarly, the asymptotic methods of stability testing, such as analyzing the eigenvalues of a Poincare map, may be inadequate for measuring a system's response to large perturbations, which can be strongly nonlinear. 

Rather, we expect a comprehensive stability analysis to involve testing across a divers distribution of failure modes. To step in that direction, we design several stability testing procedures and apply them to RL trained policies and their CPC counterparts. We have considered three types of perturbation: 1) torque noise, 2) torque modulation and 3) blows. In the first case, uncorrelated Gaussian noise is added to all the joint torques. In the second case, the joint torques are multiplicatively modulated by random factors, selected prior to a walk. In the third case, a random momentum is added to a random segment at random sparse moments of time, representing random blows to the walker. The three cases correspond to the sustained, static and punctuated perturbations respectively. In each experiment a perturbation amplitude keeps increasing until the walker falls at time $t_f$. The stability of a controller is measured by how long the walker can withstand an increasing perturbation (see accompanying video).

\subsubsection{Walker model}

\begin{figure}
  \begin{center}
    \includegraphics[width=.45\textwidth]{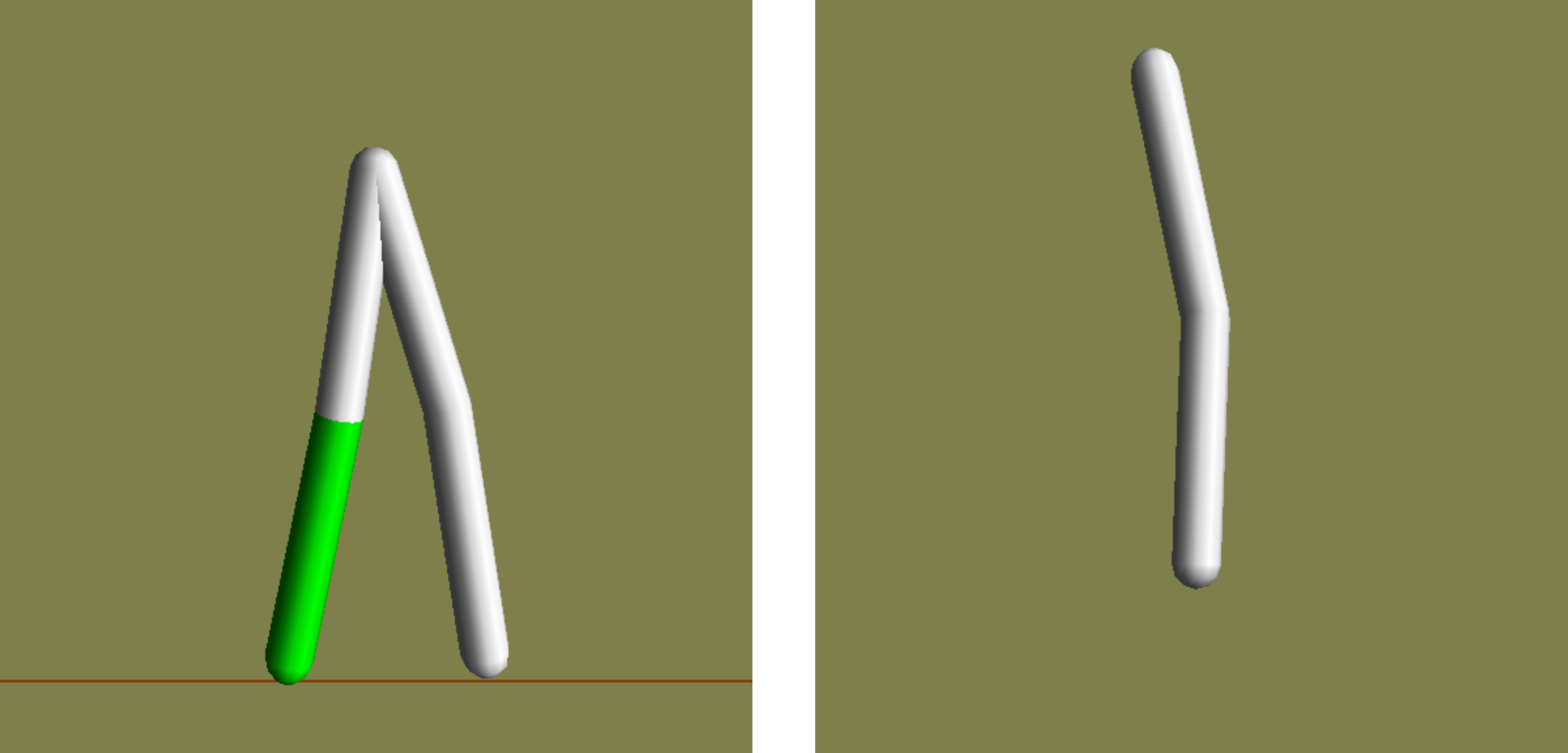}
  \end{center}
  \caption{
    Left panel: planar four-link walker model.
    Right panel: acrobot model.
  }
  \label{models}
\end{figure}

We considered one of the simplest possible kneed bipedal walkers -- a planar four-link walker with (nearly) point feet, (see Fig.(\ref{models}) and Appendix \ref{walkersetup}). The joint angles are unrestricted. It is rewarded for walking with a certain velocity and penalized for excessive control or if its COM drops too low. We trained 10 NN controllers using a variant of the natural policy gradient approach \cite{kakade2002natural}, most closely related to PPO with adaptive KL penalty \cite{schulman2017proximal}.

 To avoid the complications of dealing with different gait classes (such as backward- vs. forward-bending knees), the controller policies are initialized by an example of a humanlike-gait policy using imitation learning and random initialization of NN weights. Due to the inherent limitations of imitation learning, the initial policies are quite unstable and require substantial amount of RL training. Upon completion of training, visual inspection revealed very similar gaits in all the controllers. We verified that the mean returns and the cost of transport (COT), were within a relatively narrow range (with COT about $0.08$), suggesting that the controllers converged to a near optimal solution.

We would like to emphasize that a CPC counterpart of an RL-trained controller is constructed exclusively from a random subset $\mathbf X_d$ of the points generated by that controller (from the last RL iteration, in our implementation), withut any additional training or optimization.

\subsubsection{Results}

\begin{figure}
\includegraphics[width=.5\textwidth]{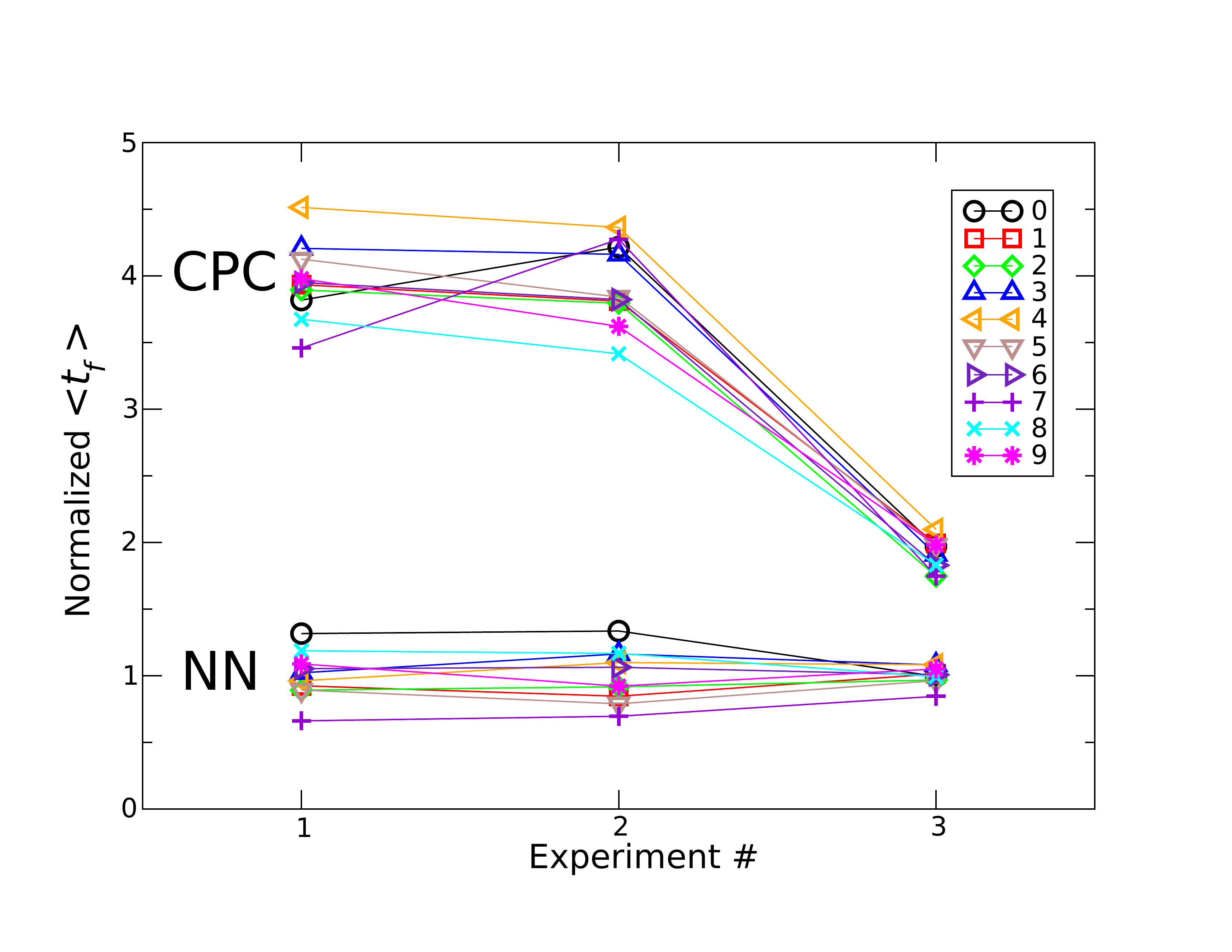}
\caption{Normalized mean fall time in three stability tests. The three tests are: 1) torque noise, 2) torque modulation, 3) blows. The top and bottom clusters are the CPC- and NN-controller results, respectively. The vertical axis indicates the improvement factor in the fall time relative to the mean NN-controller. Different symbols correspond to ten independently trained walkers.}
\label{walkertest}
\end{figure}

The stability testing results are summarized in Fig.(\ref{walkertest}). It shows $\langle t_f\rangle$ (mean $t_f$) for each controller in three different experiments. To facilitate the comparison of CPC-controllers against NN-controllers across different experiments, the results in each experiment are normalized by $t_f$ averaged over all NN-controllers in that experiment (consequently, NN controllers are centered around $\langle t_f\rangle = 1$). Each data point is the average of 100 trials. 

The plot demonstrates a substantial improvement in stability of CPC- over NN-controllers. Each CPC controller relies on $\mathbf X_d$ produced by its NN counterpart, with $|\mathbf X_d|=10^4$. The corresponding counterparts are plotted in the same color. An amplitude of perturbation in each experiment increases linearly with time, so the normalized $\langle t_f\rangle$ essentially shows the ratio of critical amplitudes (CPC over NN controllers) that make the walker fall. In terms of this ratio, the stability tests reveal a fourfold improvement in the torque noise and torque modulation experiments, and a twofold improvement in the blow experiment.

Despite the NN controllers having been trained to similar return and COT values, the variance in their stability appears quite large. Notice that the relative variance of NN controllers is somewhat larger (the factor of 2-3 in the first two tests) than the relative variance of CPC controllers. This is to be expected,  especially for the policies converged to a near optimal solution, as the properties of CPC controllers are dictated by RL training trajectories, while the stability of a RL-trained NN controller, that depends on the NN's generalization capacity, may vary significantly between individual networks.

\subsection{Acrobot}

Our second example illustrates the application of CPC to a popular class of control problems: balancing of an underactuated system around an unstable equilibrium point. In this experiment, however, no reference trajectories are provided by a functioning controller, as was the case with the planar walker. Instead, we use the CPC algorithm to construct a successful controller purely from the examples of uncontrolled failures. It is only required that the examples start at the equilibrium state, with a small random noise added to the controls to trigger a failure. To appreciate the difficulty of this task, note that it is akin to learning to balance on a tightrope simply by observing the failures of a person with no balancing skills.

The main idea is to set $s_g = -1$, which corresponds to reversing the time of reference trajectories, thus directing the system toward the equilibrium point. The idea of using a time reversal for stabilizing an unstable equilibrium is only straightforward for a fully actuated system. We investigate how well it works in practice for an underactuated system in the case of an acrobot - a double inverted pendulum with an actuated elbow, see Fig.(\ref{models}).

\begin{figure}
\includegraphics[width=.5\textwidth]{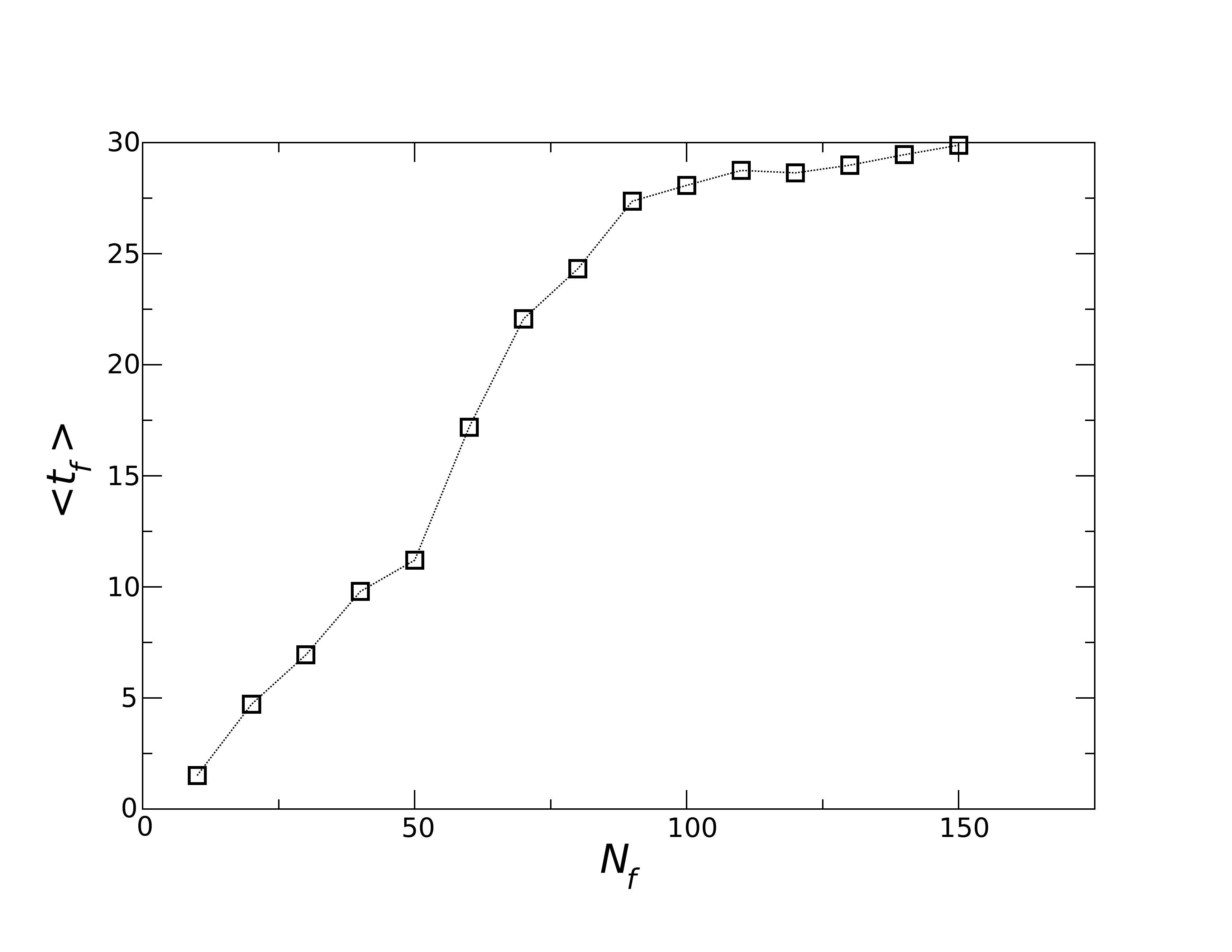}
\caption{Mean fall times $\langle t_f\rangle$ versus the number of sampled fall trajectories, tested at $6\sigma_0$ torque noise. The acrobot is observed for a maximum of 30 s, therefore $t_f\le 30$. Each data point is averaged over 100 trials, each trial includes resampling of the fall trajectories.}
\label{falltime}
\end{figure}

A CPC controller is constructed from $N_f$ fall trajectories recorded for the duration of one second ($100$ target points) starting from a completely upright configuration under a small Gaussian noise $\sigma_0$ of the joint torque, see Appendix \ref{walkersetup} for details. 
A handful of fall examples ($N_f \approx 10^1$) is sufficient to produce (with a high probability) a controller stable against the same level of the torque noise $\sigma_0$. To achieve stability against a stronger noise of $6\sigma_0$, it takes about $N_f \approx 10^2$ failure samples (corresponding to $|\mathbf X_d| = 10^4$). This is illustrated in Fig.(\ref{falltime}), displaying the mean fall time $\langle t_f \rangle$ for a given $N_f$, averaged across independent controllers, when measured to the maximum time of $T = 30$ s. Note that for large $T$ the controllers mostly fall into two categories: stable controllers with $t_f = T$ or unstable controllers with $t_f \ll T$. Therefore, $(T - \langle t_f \rangle)/T$ shows the fraction of unstable controllers. We can see that the probability of an unstable controller appears to quickly vanish as $N_f$ increases.

\section{Discussion}
\label{discussion}

In this paper we made three contributions: 1) we introduced the CPC formalism and a CPC-based algorithm for controlling underactuated systems, 2) we elucidate its relation to the ZD approach, 3) we empirically demonstrated that CPC leads to a significant enhancement of stability in RL trained policies of a bipedal walker.

We provided an explicit connection between CPC and ZD, particularly in the context of the HZD literature \cite{westervelt2018feedback}, by identifying the zero-dynamics virtual constraints that give rise to a HZD controller identical to the CPC controller in the HGL. There are substantial differences between the two approaches as well, which we summarize below, that make CPC especially suitable for stabilizing RL-learned policies.

In CPC one does not need to construct virtual constraints and to choose a gait phasing parameter, ensuring its monotonicity, as one does in HZD. This may not even be feasible for an unknown policy. Also, CPC is invariant with respect to the choice of free coordinates, while HZD is generally not. 
In practice, HZD is often used in a trajectory-centric manner, that is for periodic motion on a one-dimensional manifold. CPC, on the other hand, relies on a cloud of target points and can naturally handle a wide distribution of the initial state. 
Also, for the target point selection, CPC incorporates the expected return foresight, either from the RL value function or from the recorded data. 

The validity and utility of our approach was verified empirically by stabilizing RL-trained policies of a bipedal walker. The original policies were trained with an unbiased policy gradient method, similar to PPO. The RL policies were then compared with their CPC-stabilized counterparts in a number of stability testing experiments. In all the experiments, remarkable improvements in stability were observed, indicating a promising direction of research.

Off-line RL is an active area of research \cite{fu2020d4rl} with the potential of greatly facilitating RL, as it does not rely on expensive on-line data collection. Our approach addresses the same problem as off-line RL: converting off-line training data into a viable policy. This problem is challenging, as the off-line trained policies, with rare exception, are outperformed by the on-line trained policies \cite{fu2020d4rl}. It is all the more impressive that our approach produces superior policies greatly outperforming the original on-line RL policies in terms of stability, while relying on a small amount (only $10^4$ data points) of the off-line data. In the future, we would like to apply CPC to the standard off-line RL data sets, such as D4RL \cite{fu2020d4rl}.

\section{Acknowledgments}

We thank Georges Harik for many useful discussions.

\appendix

\section{Theorem proofs}
\label{proofs}

\subsection{Path convergence theorem}
\label{convergenceproof}

\begin{proof}
Throughout the proof we assume $0 \le t < \mathcal{O}(\varepsilon)$. Consider a system under the CPC control law $\tau_{CPC}(t)$. Let $\Delta\tau = \tau_{CPC}-\tau_d$. From Eq.(\ref{bddq}) and the definition of the renormalized target trajectory $q^r_d$ (see Eq.(\ref{qrd})) it follows that
\begin{equation}
  \Delta^r{\ddot {\bar q}} = 0 +\mathcal{O}(\Delta\tau\varepsilon,1),
  \label{bddqr}
\end{equation}
and $\chi(t)$ satisfies
\begin{equation}
  \Delta^r \ddot\chi = B_\chi \Delta \tau + \mathcal{O}(\Delta\tau\varepsilon,1).
  \label{delrddchi}
\end{equation}
Let us define: $a^{(k)}(t)$ to be the $k$-th time derivative of $a(t)$, and $z = B B_{\chi}^{-1} = [I_{M}; B_{\psi} B_{\chi}^{-1}]$. By integrating Eq.(\ref{bddqr}) one can see that, if $\Delta^r x(0)$ satisfies the reachability equation, then 
\begin{equation}
  \Delta^r {\bar q}^{(k)}(t) = 0 
  +\mathcal{O}(\Delta\tau\varepsilon^{3-k},\varepsilon^{2-k}+t_0^{2-k})
  \label{delrbqk}
\end{equation}
and 
\begin{equation}
  z \Delta^r \chi^{(k)}(t) = \Delta^r q^{(k)}(t) 
  +\mathcal{O}(\Delta\tau\varepsilon^{3-k},\varepsilon^{2-k}+t_0^{2-k})
\label{zdelrchik}
\end{equation}
for $k \in \{0,1,2\}$. Plugging in $\tau_{CPC}$ from Eq.(\ref{taucpc}) into Eq.(\ref{delrddchi}), multiplying the result by $z$ from the left and using Eq.(\ref{zdelrchik}), we find:
\begin{equation}
  \Delta^r \ddot q + \frac{k_d}{\varepsilon} \Delta^r \dot q + \frac{k_p}{\varepsilon^2} \Delta^r q = 0 
  +\mathcal{O}\left(\Delta\tau\varepsilon,1+\frac{t_0^2}{\varepsilon^2}\right).
  \label{errordynr}
\end{equation}
Note that $\Delta\tau(0) = \mathcal{O}(\Delta' q(0)/\varepsilon^2)$, and therefore $\mathcal{O}(\Delta\tau\varepsilon) = \mathcal{O}(\Delta' q(0)/\varepsilon)$. It then follows from Eq.(\ref{errordynr}), that both $\Delta^r q(t)$ and $\Delta' q(t)$ reduce down to $\mathcal{O}(\varepsilon \Delta' q(0), \varepsilon^2+t_0^2)$ on the time scale $\mathcal{O}(\varepsilon)$.
\end{proof}

\subsection{Reparameterization invariance theorem}
\label{reparameterizationproof}

\begin{proof}
We can repeat every step of the proof of Theorem \ref{convergence} till Eq.(\ref{errordynr}), but with the transformed quantities. We then arrive at:
\begin{equation}
  \Delta^r \ddot{\tilde q} + \frac{k_d}{\varepsilon} \Delta^r \dot{\tilde q} + \frac{k_p}{\varepsilon^2} \Delta^r {\tilde q} = 0
  +\mathcal{O}\left(\Delta\tau\varepsilon,1+\frac{t_0^2}{\varepsilon^2}\right).
\end{equation}
Because $C$ is invertible, we see that under $\tilde\tau_{CPC}(t)$ the system follows the exact same trajectory as under $\tau_{CPC}(t)$. Since the controls are assumed to be independent, it implies $\tilde\tau_{CPC}(t) = \tau_{CPC}(t)$.
\end{proof}

\subsection{CPC-ZD correspondence theorem}
\label{correspondenceproof}

\begin{proof}
Note that $\Delta^{ZD} q$ satisfies
\begin{equation}
  c^\T \Delta^{ZD} q^{(k)} = 0.
  \label{cdelzdqk}
\end{equation}
If $c \propto b$, then from Eqs.(\ref{delrbqk},\ref{cdelzdqk}) it follows that
\begin{equation}
  q_d^{ZD} \approx q_d^r.
\label{qdzdr}
\end{equation}
Differentiating $h(q)$ with respect to $q$ we find:
\begin{equation}
  \frac{\partial h(q)}{\partial q} = [I_M,\mathbb{0}_M] - \frac{\partial h_d(\theta(q))}{\partial \theta} c^\T,
  \label{dhdq}
\end{equation}
where $\mathbb{0}_M$ is a vector of $M$ zeros. Then from Eqs.(\ref{cdelzdqk},\ref{qdzdr},\ref{dhdq}) it follows:
\begin{multline}
  \frac{\partial h(q)}{\partial q} \Delta^r q^{(k)}
  \approx \frac{\partial h(q)}{\partial q} \Delta^{ZD} q^{(k)} 
  = \Delta^{ZD}\chi^{(k)} \\
  - \frac{\partial h_d(\theta(q))}{\partial \theta} c^\T \Delta^{ZD} q^{(k)}
  = \Delta^{ZD}\chi^{(k)} = y^{(k)} .
  \label{dhdqdelrqk}
\end{multline}
Let us define A:
\begin{equation}
  A = \frac{\partial h(q)}{\partial q} B.
  \label{A}
\end{equation}
In the HGL $B(q) \to B$ and $A(q) \to A$. We define $K_n = \left[\frac{k_p}{\varepsilon^2}, \frac{k_d}{\varepsilon}\right] \otimes I_n$. Using Eqs.(\ref{taucpc},\ref{tauzd},\ref{zdelrchik},\ref{dhdqdelrqk},\ref{A}), we can now prove the theorem:
\begin{equation}
  \begin{split}
    \Delta\tau_{CPC} = - B_{\chi}^{-1} K_M \Delta^r x_\chi
    = - A^{-1} A B_{\chi}^{-1} K_M \Delta^r x_\chi \\
    = - A^{-1} \frac{\partial h(q)}{\partial q} z K_M \Delta^r x_\chi 
    \approx - A^{-1} \frac{\partial h(q)}{\partial q} K_N \Delta^r x \\
    \approx - A^{-1}\left(\frac{k_p}{\varepsilon^2} y + \frac{k_d}{\varepsilon} \dot y\right)
    \to \Delta\tau_{ZD},
  \end{split}
\end{equation}
as $\varepsilon \to 0$.
\end{proof}

\section{Tight bounds of efficient ball-tree search}
\label{bounds}

In a ball tree, each node is associated with a sphere containing all the points of a subtree rooted at this node. Let a sphere of radius $\rho$ be centered at some point $[q_c;\dot q_c]$. A point $x_d^0$ inside the sphere can be parameterized by $(\xi,\eta)=(q_d^0-q_c, \dot q_d^0 - \dot q_c)$, with $\xi^2+\eta^2\le \rho^2$. As follows from Eqs.(\ref{t0s}, \ref{proxloss}), the proximity loss has the form:
\begin{equation}
L^{prox}=\left(\alpha_\xi+\beta_\xi^\T \xi\right)^2+\left(\alpha_\eta+\beta_\eta^\T \eta\right)^2,
\end{equation}
where 
\begin{equation}
\begin{split}
& \beta_\xi=\frac{\omega b}{\dot{\bar q}_0}, \quad \alpha_\xi=\beta_\xi^\T (q_c-q_0), \\
& \beta_\eta=\frac{b}{\dot{\bar q}_0}, \quad \alpha_\eta=\beta_\eta^\T \dot q_c - s_g .
\end{split}
\end{equation}
To construct bounds we need to only consider $\xi\propto\beta_\xi$ and $\eta\propto\beta_\eta$, because a component of $\xi$ orthogonal to $\beta_\xi$ increases $\xi^2+\eta^2$ without affecting $L^{prox}$, and similarly for $\eta$. The lowest possible loss value $L_l=0$ is reached at $(\xi,\eta)=(\xi_0,\eta_0)$
\begin{equation}
(\xi_0,\eta_0)=\left(
-\frac{\alpha_\xi \beta_\xi}{\lVert\beta_\xi\rVert^2},
-\frac{\alpha_\eta \beta_\eta}{\lVert\beta_\eta\rVert^2}
\right),
\end{equation}
provided $\xi_0^2+\eta_0^2\le \rho^2$. Otherwise, the lower bound $L_l>0$ and it should be looked for on the sphere $\xi^2+\eta^2=\rho^2$. In that case we parameterize $\xi$ and $\eta$ as:
\begin{equation}
(\xi,\eta)=\left(
\frac{\beta_\xi}{\lVert\beta_\xi\rVert}\rho\cos{\theta},
\frac{\beta_\eta}{\lVert\beta_\eta\rVert}\rho\sin{\theta}
\right) .
\end{equation}
The extrema of $L^{prox}$ are given by $\partial{L^{prox}}/{\partial\theta}=0$:
\begin{multline}
\rho\left(\lVert\beta_\xi\rVert^2-\lVert\beta_\eta\rVert^2\right)\cos{\theta}\sin{\theta}
+ \alpha_\xi \lVert\beta_\xi\rVert \sin{\theta} \\
= \alpha_\eta \lVert\beta_\eta\rVert \cos{\theta}.
\label{extremacond}
\end{multline}
Squaring both sides of Eq.(\ref{extremacond}) we obtain a fourth degree polynomial in $\cos{\theta}$
\begin{multline}
\left[\rho\left(\lVert\beta_\xi\rVert^2-\lVert\beta_\eta\rVert^2\right)\cos{\theta} 
+ \alpha_\xi \lVert\beta_\xi\rVert\right]^2(1-\cos^2{\theta}) \\
- \left(\alpha_\eta \lVert\beta_\eta\rVert \cos{\theta}\right)^2 = 0,
\end{multline}
that can be solved analytically. The lowest and highest values of the loss corresponding to the real roots $\in[-1,1]$ of the polynomial, that also satisfy Eq.(\ref{extremacond}), are then the lower and upper bounds $L_l$ and $L_u$.

\section{Details of the experimental setup}
\label{walkersetup}

Our experiments were simulated using Open Dynamics Engine (ODE). The walker is modeled as a planar chain of 4 identical segments (capped ODE cylinders of length $1$  radius $0.1$ and density $1$), connected by powered joints. We set the gravitational acceleration to $10$, the simulation time step $\Delta t$ to $0.005$ and the ODE friction coefficient parameter to $1$. The control matrix $B$ is estimated from the last $n = 9$ points of the current trajectory. 

 We have not searched extensively for the algorithm hyperparameter values, settling instead with the values that appeared reasonable after a few trials: $\omega=10$, $s_g=1$, $n_d=20$, $k_0=2000$, $\tau_c=2$ and $k_c=2$.

 Within the RL framework, the walker is controlled by a Gaussian policy $\tau \sim \mathcal{N}(\mu(x),\sigma_0^2)$ with fixed noise strength $\sigma_0=0.02$ and $\mu(x)$ represented by a 5-layer NN with 4860 weights. The walker's goal is to walk efficiently with a velocity close to $v_g=1$. To increase the gait stability, the reward function includes a penalty term for swinging a foot too close to the ground.

In the acrobot case, the parameters that differ from the walker setup are: $s_g = -1$, $\Delta t = 0.01$ and $n = 7$. Also, we set $\tau_d = V(x_d) = r(x_d) = 0$ for the purpose of computing $V_I$ and $V_{II}$ in Eqs.(\ref{vi},\ref{vii}), as no good reference policy is provided in this task.

\bibliographystyle{plain}
\bibliography{cpc_arxiv.bbl}

\begin{thebibliography}{10}

\bibitem{anderson2007optimal}
Brian~DO Anderson and John~B Moore.
\newblock {\em Optimal control: linear quadratic methods}.
\newblock Courier Corporation, 2007.

\bibitem{bledt2018cheetah}
Gerardo Bledt, Matthew~J Powell, Benjamin Katz, Jared Di~Carlo, Patrick~M
  Wensing, and Sangbae Kim.
\newblock Mit cheetah 3: Design and control of a robust, dynamic quadruped
  robot.
\newblock In {\em 2018 IEEE/RSJ International Conference on Intelligent Robots
  and Systems (IROS)}, pages 2245--2252. IEEE, 2018.

\bibitem{byrnes1991asymptotic}
Christopher~I Byrnes and Alberto Isidori.
\newblock Asymptotic stabilization of minimum phase nonlinear systems.
\newblock {\em IEEE Transactions on Automatic Control}, 36(10):1122--1137,
  1991.

\bibitem{duan2016benchmarking}
Yan Duan, Xi~Chen, Rein Houthooft, John Schulman, and Pieter Abbeel.
\newblock Benchmarking deep reinforcement learning for continuous control.
\newblock In {\em International Conference on Machine Learning}, pages
  1329--1338, 2016.

\bibitem{fu2020d4rl}
Justin Fu, Aviral Kumar, Ofir Nachum, George Tucker, and Sergey Levine.
\newblock D4rl: Datasets for deep data-driven reinforcement learning.
\newblock {\em arXiv preprint arXiv:2004.07219}, 2020.

\bibitem{gibney2016google}
Elizabeth Gibney.
\newblock Google ai algorithm masters ancient game of go.
\newblock {\em Nature News}, 529(7587):445, 2016.

\bibitem{gong2019feedback}
Yukai Gong, Ross Hartley, Xingye Da, Ayonga Hereid, Omar Harib, Jiunn-Kai
  Huang, and Jessy Grizzle.
\newblock Feedback control of a cassie bipedal robot: Walking, standing, and
  riding a segway.
\newblock In {\em 2019 American Control Conference (ACC)}, pages 4559--4566.
  IEEE, 2019.

\bibitem{griffin2017nonholonomic}
Brent Griffin and Jessy Grizzle.
\newblock Nonholonomic virtual constraints and gait optimization for robust
  walking control.
\newblock {\em The International Journal of Robotics Research}, 36(8):895--922,
  2017.

\bibitem{grizzle2017virtual}
Jessy~W Grizzle and Christine Chevallereau.
\newblock Virtual constraints and hybrid zero dynamics for realizing
  underactuated bipedal locomotion.
\newblock {\em arXiv preprint arXiv:1706.01127}, 2017.

\bibitem{gu2017deep}
Shixiang Gu, Ethan Holly, Timothy Lillicrap, and Sergey Levine.
\newblock Deep reinforcement learning for robotic manipulation with
  asynchronous off-policy updates.
\newblock In {\em Robotics and Automation (ICRA), 2017 IEEE International
  Conference on}, pages 3389--3396. IEEE, 2017.

\bibitem{hanna2017grounded}
Josiah~P Hanna and Peter Stone.
\newblock Grounded action transformation for robot learning in simulation.
\newblock In {\em AAAI}, pages 3834--3840, 2017.

\bibitem{hinton12deep}
Geoffrey Hinton, Li~Deng, Dong Yu, George~E Dahl, Abdel-rahman Mohamed, Navdeep
  Jaitly, Andrew Senior, Vincent Vanhoucke, Patrick Nguyen, Tara~N Sainath,
  et~al.
\newblock Deep neural networks for acoustic modeling in speech recognition: The
  shared views of four research groups.
\newblock {\em IEEE Signal Processing Magazine}, 29(6):82--97, 2012.

\bibitem{horn2018hybrid}
Jonathan~C Horn, Alireza Mohammadi, Kaveh~Akbari Hamed, and Robert~D Gregg.
\newblock Hybrid zero dynamics of bipedal robots under nonholonomic virtual
  constraints.
\newblock {\em IEEE Control Systems Letters}, 3(2):386--391, 2018.

\bibitem{hutter2016anymal}
Marco Hutter, Christian Gehring, Dominic Jud, Andreas Lauber, C~Dario
  Bellicoso, Vassilios Tsounis, Jemin Hwangbo, Karen Bodie, Peter Fankhauser,
  Michael Bloesch, et~al.
\newblock Anymal-a highly mobile and dynamic quadrupedal robot.
\newblock In {\em 2016 IEEE/RSJ International Conference on Intelligent Robots
  and Systems (IROS)}, pages 38--44. IEEE, 2016.

\bibitem{james2017transferring}
Stephen James, Andrew~J Davison, and Edward Johns.
\newblock Transferring end-to-end visuomotor control from simulation to real
  world for a multi-stage task.
\newblock In {\em Conference on Robot Learning}, pages 334--343. PMLR, 2017.

\bibitem{kakade2002natural}
Sham~M Kakade.
\newblock A natural policy gradient.
\newblock In {\em Advances in neural information processing systems}, pages
  1531--1538, 2002.

\bibitem{kober2013reinforcement}
Jens Kober, J~Andrew Bagnell, and Jan Peters.
\newblock Reinforcement learning in robotics: A survey.
\newblock {\em The International Journal of Robotics Research},
  32(11):1238--1274, 2013.

\bibitem{krizhevsky2012imagenet}
Alex Krizhevsky, Ilya Sutskever, and Geoffrey~E Hinton.
\newblock Imagenet classification with deep convolutional neural networks.
\newblock In {\em Advances in neural information processing systems}, pages
  1097--1105, 2012.

\bibitem{kuindersma16}
Scott Kuindersma, Robin Deits, Maurice~F. Fallon, Andres Valenzuela, Hongkai
  Dai, Frank Permenter, Twan Koolen, Pat Marion, and Russ Tedrake.
\newblock Optimization-based locomotion planning, estimation, and control
  design for the atlas humanoid robot.
\newblock {\em Auton. Robots}, 40(3):429--455, 2016.

\bibitem{lecun15deep}
Yann LeCun, Yoshua Bengio, and Geoffrey Hinton.
\newblock Deep learning.
\newblock {\em nature}, 521(7553):436, 2015.

\bibitem{levine2018learning}
Sergey Levine, Peter Pastor, Alex Krizhevsky, Julian Ibarz, and Deirdre
  Quillen.
\newblock Learning hand-eye coordination for robotic grasping with deep
  learning and large-scale data collection.
\newblock {\em The International Journal of Robotics Research},
  37(4-5):421--436, 2018.

\bibitem{mandlekar2017adversarially}
Ajay Mandlekar, Yuke Zhu, Animesh Garg, Li~Fei-Fei, and Silvio Savarese.
\newblock Adversarially robust policy learning: Active construction of
  physically-plausible perturbations.
\newblock In {\em 2017 IEEE/RSJ International Conference on Intelligent Robots
  and Systems (IROS)}, pages 3932--3939. IEEE, 2017.

\bibitem{mnih2013playing}
Volodymyr Mnih, Koray Kavukcuoglu, David Silver, Alex Graves, Ioannis
  Antonoglou, Daan Wierstra, and Martin Riedmiller.
\newblock Playing atari with deep reinforcement learning.
\newblock {\em arXiv preprint arXiv:1312.5602}, 2013.

\bibitem{muratore2019assessing}
Fabio Muratore, Michael Gienger, and Jan Peters.
\newblock Assessing transferability from simulation to reality for
  reinforcement learning.
\newblock {\em IEEE Transactions on Pattern Analysis and Machine Intelligence},
  2019.

\bibitem{pinto2017robust}
Lerrel Pinto, James Davidson, Rahul Sukthankar, and Abhinav Gupta.
\newblock Robust adversarial reinforcement learning.
\newblock In {\em International Conference on Machine Learning}, pages
  2817--2826. PMLR, 2017.

\bibitem{raibert2008bigdog}
Marc Raibert, Kevin Blankespoor, Gabriel Nelson, and Rob Playter.
\newblock Bigdog, the rough-terrain quadruped robot.
\newblock {\em IFAC Proceedings Volumes}, 41(2):10822--10825, 2008.

\bibitem{sadeghi2016cad2rl}
Fereshteh Sadeghi and Sergey Levine.
\newblock {CAD2RL:} real single-image flight without a single real image.
\newblock In {\em Robotics: Science and Systems XIII}, 2017.

\bibitem{schulman2015trust}
John Schulman, Sergey Levine, Pieter Abbeel, Michael Jordan, and Philipp
  Moritz.
\newblock Trust region policy optimization.
\newblock In {\em International Conference on Machine Learning}, pages
  1889--1897, 2015.

\bibitem{schulman2015high}
John Schulman, Philipp Moritz, Sergey Levine, Michael~I. Jordan, and Pieter
  Abbeel.
\newblock High-dimensional continuous control using generalized advantage
  estimation.
\newblock In {\em 4th International Conference on Learning Representations},
  2016.

\bibitem{schulman2017proximal}
John Schulman, Filip Wolski, Prafulla Dhariwal, Alec Radford, and Oleg Klimov.
\newblock Proximal policy optimization algorithms.
\newblock {\em arXiv preprint arXiv:1707.06347}, 2017.

\bibitem{shigemi2018asimo}
Satoshi Shigemi, A~Goswami, and P~Vadakkepat.
\newblock Asimo and humanoid robot research at honda.
\newblock {\em Humanoid robotics: A reference}, pages 1--36, 2018.

\bibitem{singh2017machine}
Shashi~Pal Singh, Ajai Kumar, Hemant Darbari, Lenali Singh, Anshika Rastogi,
  and Shikha Jain.
\newblock Machine translation using deep learning: An overview.
\newblock In {\em 2017 International Conference on Computer, Communications and
  Electronics (Comptelix)}, pages 162--167. IEEE, 2017.

\bibitem{tobin2017domain}
Josh Tobin, Rachel Fong, Alex Ray, Jonas Schneider, Wojciech Zaremba, and
  Pieter Abbeel.
\newblock Domain randomization for transferring deep neural networks from
  simulation to the real world.
\newblock In {\em Intelligent Robots and Systems (IROS), 2017 IEEE/RSJ
  International Conference on}, pages 23--30. IEEE, 2017.

\bibitem{venkataraman2019recovering}
Harish~K Venkataraman and Peter~J Seiler.
\newblock Recovering robustness in model-free reinforcement learning.
\newblock In {\em 2019 American Control Conference (ACC)}, pages 4210--4216.
  IEEE, 2019.

\bibitem{wang2010optimizing}
Jack~M Wang, David~J Fleet, and Aaron Hertzmann.
\newblock Optimizing walking controllers for uncertain inputs and environments.
\newblock {\em ACM Transactions on Graphics (TOG)}, 29(4):1--8, 2010.

\bibitem{westervelt2018feedback}
Eric~R Westervelt, Jessy~W Grizzle, Christine Chevallereau, Jun~Ho Choi, and
  Benjamin Morris.
\newblock {\em Feedback control of dynamic bipedal robot locomotion}.
\newblock CRC press, 2018.

\bibitem{young2018recent}
Tom Young, Devamanyu Hazarika, Soujanya Poria, and Erik Cambria.
\newblock Recent trends in deep learning based natural language processing.
\newblock {\em ieee Computational intelligenCe magazine}, 13(3):55--75, 2018.

\end{thebibliography}

\end{document}